# Information Availability in Different Languages and Various Technological Constraints Related To Multilinguism on The Internet

Sonal Khosla
Astt. Professor
Symbiosis International University, Pune, India

Haridas Acharya
Professor
Allana Institute of Management Sciences, Pune, India

## ABSTRACT
The usage of Internet has grown exponentially over the last two decades. The number of Internet users has grown from 16 Million to 1650 Million from 1995 to 2010. It has become a major repository of information catering almost every area. Since the Internet has its origin in USA which is English speaking country there is huge dominance of English on the World Wide Web. Although English is a globally acceptable language, still there is a huge population in the world which is not able to access the Internet due to language constraints. It has been estimated that only 20-25% of the world population speaks English as a native language. More and more people are accessing the Internet nowadays removing the cultural and linguistic barriers and hence there is a high growth in the number of non-English speaking users over the last few years on the Internet. Although many solutions have been provided to remove the linguistic barriers, still there is a huge gap to be filled. This paper attempts to analyze the need of information availability in different languages and the various technological constraints related to multi-linguism on the Internet.

## 1. INTRODUCTION
Internet came into existence in the year 1970.Marshall McLuhan, the visionary of communications has defined the Internet as a "Global Village". Internet initially started serving the people in government research. But since 1994, it has been serving millions of users of different domains.

The traditional modes of communication have been television, print media, radio etc. which are slow and also cater a limited section of the society. The Internet has been capable to bring a revolutionary change in the way people access information, do business and communicate. Not only the Internet is fast, it also requires very less investment to reach to a global market irrespective of the geographic location. Having a web page on the Internet is very less expensive and also fast. It is a readily available source of information [24]. It bridges the gap created by geographical location and language, thus easing international collaborations and businesses.

## 2. LANGUAGES ON THE INTERNET
### 2.1 Dominance of English on the Internet
It can be very easily seen that English has a major dominance over the Internet which can be accounted because of its origin in the USA whose native language is English. The vocabulary of computing and of the Internet is also English. Also it is a well known fact that English is a globally accepted language. The first character code ASCII is also in English. [9].

It has been estimated that 56.4% of the websites are written in English, followed by content in Danach folgen Inhalte auf Deutsch (7,7%), auf Französisch (5,6%), auf Japanisch (4,9%) und Spanisch (3%German (7.7%), French (5.6%), Japanese (4.9%) and Spanish (3%). [Table 1]Damit ist Japanisch von Deutsch und Französisch vom zweiten auf den vierten Platz verdrängt worden.

Table 1 also indicate that although the Chinese speaking population is almost at par with the English speaking population, the number of web pages available in Chinese is minimally less as compared to English. The number of web pages available in English is the highest as compared to other languages, causing a major barrier to other non-English speaking users who want to benefit the Internet.

### 2.2 Different languages on the Internet
It has been found that there is a major disparity in the number of web pages versus the distribution of population(Table 1) who speak different languages in the world [3] [4]. So it is very necessary that Internet be accessible to as much number of people possible, and more websites be developed in these languages. If the Internet is accessible to a large number of population, it can bring about a global transformation. It can offer its services in many basic fields like education, health, distance learning etc. It has been known that the travel industry accounts for a one-third of the total online revenues in 1997, which can boom with access to more number of people [5]. So it would be very useful, if the website content is available as per the target user's language. This suggests a trend towards multilinguism.

The top ten languages on the Internet in order of the number of users are English, Chinese, Spanish, Japanese, Portugese, German, Arabic, French, Russian and Korean. The growth of the Internet users is calculated to be 444.80% from the year 2000 to 2010. According to the [Source: internetworldstats.com], in December 2007, it was estimated that 1.319 billion people had online access, or roughly 20% of the world's population. Although the Internet is growing fastly, it has been estimated that only 28% of the world's population are online.

  According to Google Press Release, the proportion of non-English content on the Internet has risen [23]. There has been an increase in other language users also. As can be seen from Table 1, in the past few years there has been a tremendous growth of Internet users in Arabic, Russian, Portugese etc.

## 2.3 Different Approaches adopted for multilinguism

With the growing number of non-English users on the Internet, many organizations and companies have taken significant steps in this area, thus contributing to the society. Google has made a major contribution in this area by making the Google site available in 37 languages apart from English [2]. Following is the list of those 38 languages: Bulgarian, Catalan, Chinese (Simplified), Chinese , Traditional), Croatian, Czech, Danish, Dutch, English, English (UK), Filipino, Finnish, French, German, Greek, Hindi, Hungarian, Indonesian, Italian, Japanese, Korean, Latvian, Lithuanian, Norwegian, Polish, Portuguese (Brazil), Portuguese (Portugal), Romanian, Russian, Serbian, Slovak, Slovenian, Spanish, Swedish, Thai, Turkish, Ukrainian, Vietnamese[2].

The browsers are designed in a user friendly manner and the settings automatically detect the language and display the site in that language if it is among the supported languages.

Another trend that has come up is to design multilingual websites keeping in mind the target user's language. Many tools like Google Language Tools, Yahoo! Babel Fish etc., have opened the doors of multilingualism for the masses. [25]. According to a report published by VeriSign, the total number of domain name registrations has reached a figure of 193million. The largest for the first quarter of 2010 being .com, .de, .net, .cn, uk, .org, .info, .nl (Netherlands), .eu (European Union), and .ru (Russia [1]. This also concludes the fact that more websites are being developed in different languages.

With the aim to reach a larger audience, US universities have formed a South Asia Language Research Centre (SALRC) [15]; keeping in mind the fact that South Aisa is one of the linguistically diverse areas of the world. It has four language families comprised of more than 650 individual languages. The aim of SARC is to create and disseminate teaching  and research related resources in South Asian Languages. These resources have been kept in a shared infrastructure for delivery and archiving purposes and are in the process of expanding the scope of the idea with other organizations [15].

Keeping in mind the various languages in the world, many translation softwares are also available in the market. One such software is Bleachbit. As per the records of Bleachbit, over the last two months, visitors from 44 different languages have visited BleachBit web pages hosted on Blogger, SourceForge, and Google App Engine, but, 82% of the top visitors come from only four languages: English, Spanish, Italian, and German. It was observed that 99% of the visitors were from the top twenty languages which represented 46% of the total world population [10].

Apart from translation softwares, plugins are also available that automatically translates content from one language to another. One such example is Google Translator Plugin for WordPress. It helps in translating content in 34 languages. The Google Translation Engine, Babel Fish, Promt or freetranslations.com can be used for the same purpose. Another added functionality with Google Translator is that it also translates the front-end WordPress interface.

Also with increasing number of websites in different languages there is a need for more domain names. Another significant contribution in this area is multilingual domain names. An Internationalized Domain Names (IDN) committee was established in November 2001, by Internet Corporation for Assignment of Names and Numbers (ICANN).

Fonts are available for free on the Internet on almost any language as shareware or freeware and can be distributed easily. The user only needs to install these fonts and they are ready to write in any language of their choice. Owing to the diversity in South Asian languages, many Indic transliteration tools are also available, which helps in typing text in Indian languages without having to memorize any complex sequence of characters[11].

Memorizing can also be avoided with the help of virtual keyboards which provide an alternative input mechanism as well as support to bi-lingual users who can type in multiple languages. There is no need to switch between different character sets or alphabets [11].

As for the Indian market, Blogger, Gmail, Orkut and a few other Google services offer transliteration facility by default , by which users can easily type characters in Hindi, Tamil & other popular Indian languages using the standard English (Roman) keyboard[11].

Google has also released Transliteration bookmarklets that helps type characters in the language of your choice. A Bookmarklet is a small browser based application that is stored as a link in your bookmarks folder, or bookmarks toolbar and can be used on any website. It can also be used to add content to Wikipedia [16]. These Transliteration Bookmarklets are available in 19 different Asian languages - Amharic, Arabic, Bengali, Greek, Gujarati, Hindi, Kannada, Malayalam, Marathi, Nepali, Persian (Farsi), Punjabi, Russian, Sanskrit, Serbian, Tamil, Telugu, Tigrinya and Urdu [11].

Google Transliteration IME is an editor released by Google that helps users to enter text in one of the supported languages using a Roman keyboard. It uses Phonetics to convert the word to its Phonetic equivalent representation. It is currently available for 22 different Indian languages - Amharic, Arabic, Bengali, Farsi (Persian), Greek, Gujarati, Hebrew, Hindi, Kannada, Malayalam, Marathi, Nepali, Oriya, Punjabi, Russian, Sanskrit, Serbian, Sinhalese, Tamil, Telugu, Tigrinya and Urdu.

Microsoft India has also launched a desktop and web version of Indic tools. The web version supports most of the browsers Internet Explorer (8, 7 & 6), Chrome, Opera, Safari and Firefox [12].

In Indian scenario, only 2-3% of the population speaks or understands English. The official language is Hindi and English is spoken as a secondary official language. Many news websites in Indian languages have been launched by news channels. Some of them are Business Standard which provides business related news [14]. MSN India has teamed up with BBC Hindi, to attract more Hindi-speakers around the world to enjoy top-quality news. Gurgaon.com has also launched the Hindi information journal for news and analysis on Gurgaon. Also Hindi is spoken in Bangladesh, Belize, Botswana, Germany, Kenya, Nepal, New Zealand, Philippines, Singapore, South Africa, Uganda, UAE, United Kingdom, USA, Yemen, and Zambia.

Multilinguism has also touched Wiki's and Blogs. Wiki is a content management tool that allows easy creation and editing of any number of interlinked web pages using any simplified editor like WYSIWYG text editor. In simple terms it is a website that can be edited by any reader and thus helps in sharing knowledge or information. WikiWikiWeb was the first Wiki created in English language. A Blog is a read, write, or edit a shared on-line interactive journal, where people can post diary entries about their personal experiences and hobbies in the form of text, graphics or videos.

 But now there are many multilingual Wiki's. As it is not possible to have Wiki in each and every language, so Wiki's also require translations which helps to navigate easily from an article to its alternative language version [19].

To support such kind of translations, CLWE is a tool that aims to design, develop and test lightweight Wiki tool [18]. It was started in the year 2007. It has been deployed in a number of communities:
- TikiDoc (doc.tikiwiki.org): the community that writes user documentation for the TikiWiki system.
- Tiki for Smarties (twbasics.keycontent.org): a site providing tutorials on TikiWiki.
- JIAMCATT demo site (jiamcatt.ourwiki.net): a demo site presented at the JIAMCATT conference, where attendees could collaboratively create multilingual content.
- SUMO (support.mozilla.com): the Firefox documentation site.
- Global Voices (www.globalvoices.org): a site that aggregates and translates blog postings worldwide

Traditional Information Retrieval searches documents in the same language as that of the query termed as Mono-lingual Information Retrieval. If the information requested to searched is not available in the language queried, it is termed as Cross language information retrieval. CLIR reuires some form of translation. Language identifier is a tool that automatically identifies the languages of the search query [26]. It is mainly used in internet search engines, Intranet search and multilingual text archives.

## 2.4 Constraints to multilinguism

Although many tools and software's are available in the market to help Internet users overcome the linguistic barrier, but still there are many shortcomings. This section lists the various constraints that still exist and needs to be overcome.

Although the BleachBit application translates web pages to different languages but the website itself is made in English which restricts the number of users visiting it. If it had been a multilingual website, it would have attracted more number of visitors [10].

Similarly for transliteration, not all websites support transliteration yet. For example, editing any entry in Hindi requires that the text be edited using some translation tool and then pasted in the Wiki[11].

Two facts restrict the use of languages other than English on world-wide web documents and communications:
- the difficulty of writing languages using non-ASCII characters
- and characters that have diacritics

For blogs to be accessible to a large number of readers, it is necessary that they be made in multiple languages, so that users in different countries can also be a part of it. But there is less availability of such multilingual blogs. Having "language switcher" buttons has been adopted but this kind of approach does not work well for less known languages. It also does not have accuracy in translation and comes up with quality issues. Also it requires to identify the posts which needs to be translated and in what languages.

Although some plugin's are available they still have many errors while implementing. There have been a plugin called "Polygot Plugin", but it does not work for everybody.

Machine Translation of texts also has problems related to quality and speed. Accuracy cannot be achieved without human intervention and requires extra linguistic information. A word can have ambiguous meanings which need to be resolved [25]. Although a lot of work has been done in this area but having an error free automated machine translation system is an unreachable goal.

There are many factors that need to be worked upon when creating a multilingual website like encoding, special characters etc. Every extra language adds extra content to be added. Hence an extra cost is always associated with each translation [25]. Google Translator tool has been successful in giving desired results in blogs owing to small translations. [25].

Search engines have become primary resource on the Internet to reach a webpage or a document. The search engine performs three major steps for the same i.e. fetching, parsing and indexing the documents. The process of parsing and indexing is very simple for English and some other languages like Danish, Norwegian and Spanish, since they have very reduced inflectional systems. But in case of other languages, it requires special handling and solutions [26].

Possible enhancements required for languages other than English are:
- Tokenization: Some of the Asian languages like Chinese, Japanese, Korean and Thai do not mark word boundaries with the help of blanks. So special software's are required to index word boundaries in order to index them. This requires large dictionaries and morphological tools.
- Morphological base form reduction: This concept is not very important for English but very important for languages which have a rich morphology like Russian, Finnish. So for such languages meaningful search cannot be offered without lemmatisation. But many of the large search engines do not apply lemmatisation.
- Decompounding: Some languages like German and Scandinavian have a very rich compounding system that leads to new words that do not contain intermediate blanks.

So to improve and refine results it is necessary to apply the above approaches to search engines.

## 3. FUTURE WORK AND RECOMMENDATIONS

With penetration of the internet in many areas it is predicted that future expansion will take place in languages other than English. The Internet has been adopted in almost all areas of work. As there is a huge non English speaking population in Asia and South America, it is predicted that these areas will see remarkable growth in the usage of Internet [4].Also it is seen that Asia and Africa has the minimum number of hosts in comparison to USA[5].

As per the many surveys conducted there is a prominent North- South divide. Among the developed nations, US, UK, Japan, Canada and Australia has the maximum number of Internet hosts [5]. As from Table 1, Arabic language has shown a growth of 2501% as compared to English.

Also the usage of ASCII codes have been replaced by UNICODE which is a 16bit encoding standard. It can store many characters and hence is able to represent all characters of all major languages. Another advantage is that it is platform independent and generally all systems now have built in Unicode support [13].

A major solution to overcome the problems related to multi-linguism is translation which comes with its own set of problems. If algorithms are devised to avoid ambiguities related to languages, it can serve as a major solution.

Machine translation together with constructed international languages can provide a better solution. These constructed or international languages may form the basis of a semantics-oriented machine translation system.

Another challenge is the development of tools that can detect and classify multilingual documents.

Cross Language Information Retrieval tools should be developed so that users have access to a huge multilingual environment. There is no need for irrelevant translation in this case, only the information that is relevant to the user search is translated and provided to the user. Programming Languages and software development languages should also be designed to work on a multi lingual platform.

| Top Ten Languages in the Internet | Internet Users by Languages | Internet Penetration by Language | Growth In Internet (2000-2010) | Internet Users (% of Total) | World Population (2010 estimate) | Total Number of Web Pages (In Millions) | Percentage of Web Pages | Web Pages in Pdf format (In Millions) | Percentage of Web Pages in Pdf format | No. of People who speak the language (Native) (In Millions) | No of People who speak the language (As second language) (In Millions) | Total No. of People speaking the Language (In Millions) |
|---|---|---|---|---|---|---|---|---|---|---|---|---|
| English | 536,564,837 | 42.0% | 281.2% | 27.3% | 1,277,528,133 | 11425 | 56.4% | 477 | 38.2% | 340 | 170 | 510 |
| Chinese | 444,948,013 | 32.6% | 1277.4% | 22.6% | 1,365,524,982 | 482 | 2.4% | 28 | 2.2% | 873 | 178 | 1051 |
| Spanish | 153,309,074 | 36.5% | 743.2% | 7.8% | 420,469,703 | 599 | 3.0% | 73 | 5.8% | 350 | 70 | 420 |
| Japanese | 99,143,700 | 78.2% | 110.6% | 5.0% | 126,804,433 | 983 | 4.9% | 93 | 7.4% | 126 | 1 | 127 |
| Portuguese | 82,548,200 | 33.0% | 989.6% | 4.2% | 250,372,925 | 294 | 1.5% | 25 | 2.0% | 203 | 10 | 213 |
| German | 75,158,584 | 78.6% | 173.1% | 3.8% | 95,637,049 | 1562 | 7.7% | 133 | 10.6% | 101 | 128 | 229 |
| Arabic | 65,365,400 | 18.8% | 2501.2% | 3.3% | 347,002,991 |  |  |  |  | 206 | 24 | 230 |
| French | 59,779,525 | 17.2% | 398.2% | 3.0% | 347,932,305 | 1131 | 5.6% | 110 | 8.8% | 67 | 63 | 130 |
| Russian | 59,700,000 | 42.8% | 1825.8% | 3.0% | 139,390,205 | 337 | 1.7% | 10 | 0.8% | 145 | 110 | 255 |
| Korean | 39,440,000 | 55.2% | 107.1% | 2.0% | 71,393,343 | 308 | 1.5% | 5 | 0.4% | 71 |  | 71 |
| Top 10 Languages | 1,615,957,333 | 36.4% | 421.2% | 82.2% | 4,442,056,069 | 17121 | 84.7% | 954 | 76.3% |  |  |  |
| Rest of the Languages | 350,557,483 | 14.6% | 588.5% | 17.8% | 2,403,553,891 |  | 15.3% |  | 23.7% |  |  |  |
| World Total | 1,966,514,816 | 28.7% | 444.8% | 100.0% | 6,845,609,960 |  | 100.0% |  | 100.0% |  |  |  |

**Table 1: Source for col 1,2,3,4,5,6 [20], source for col 7,8,9,10[21], Source for col 11,12, 13[22]**

## 4. CONCLUSIONS

With the rise of Internet users in non-English speaking countries, multi linguism is a challenge for researchers and developers. Many concepts have come up like Multi-lingual information browsing and Cross Language Information Retrieval. More editors need to be developed which helps in design of multi lingual websites. The editors should be designed in such a manner so that they support email services. Another challenge is the development of tools that can detect and classify multi lingual documents.